\documentclass[letterpaper, 10 pt, conference]{ieeeconf}  

\IEEEoverridecommandlockouts                              

\usepackage{etoolbox}
\makeatletter
\patchcmd{\@makecaption}
  {\scshape}
  {}
  {}
  {}

\let\endtable\relax
\def\endtable{\end@float}
\let\endfigure\relax
\def\endfigure{\end@float}
\makeatother
\usepackage[caption=false, font=footnotesize]{subfig}
\usepackage{amsfonts}
\usepackage{multirow}
\usepackage{threeparttable}
\usepackage{pifont}
\usepackage{graphicx}  
\usepackage{booktabs}
\usepackage{authblk}
\usepackage[T1]{fontenc}
\newcommand{\cmark}{\ding{51}}
\newcommand{\xmark}{\ding{55}}

\makeatletter
\renewcommand\AB@affilsepx{\quad \protect\Affilfont} 
\makeatother

\title{\LARGE \bf
Long-Term Memory for VLA-based Agents \\ in Open-World Task Execution
}

\author[1, *]{Xu Huang}
\author[2, *]{Weixin Mao}
\author[2]{Yinhao Li}
\author[1, \textdagger]{Jiabao Zhao}
\author[2]{Hua Chen}

\affil[ ]{*Equal Contribution; \textdagger Corresponding author \authorcr }

\affil[1]{Nanjing University \authorcr}
\affil[2]{LimX Dynamics}

\begin{document}

\maketitle

\thispagestyle{empty}
\pagestyle{empty}

\begin{abstract}
Vision-Language-Action (VLA) models have demonstrated significant potential for embodied decision-making; however, their application in complex chemical laboratory automation remains restricted by limited long-horizon reasoning and the absence of persistent experience accumulation. Existing frameworks typically treat planning and execution as decoupled processes, often failing to consolidate successful strategies, which results in inefficient trial-and-error in multi-stage protocols. In this paper, we propose \textbf{ChemBot}, a dual-layer, closed-loop framework that integrates an autonomous AI agent with a progress-aware VLA model (\textbf{Skill-VLA}) for hierarchical task decomposition and execution. ChemBot utilizes a dual-layer memory architecture to consolidate successful trajectories into retrievable assets, while a Model Context Protocol (MCP) server facilitates efficient sub-agent and tool orchestration. To address the inherent limitations of VLA models, we further implement a future-state-based asynchronous inference mechanism to mitigate trajectory discontinuities. Extensive experiments on collaborative robots demonstrate that ChemBot achieves superior operational safety, precision, and task success rates compared to existing VLA baselines in complex, long-horizon chemical experimentation.
\end{abstract}

\section{INTRODUCTION}

\begin{figure}[htbp]
    \centering
    \includegraphics[width=\linewidth]{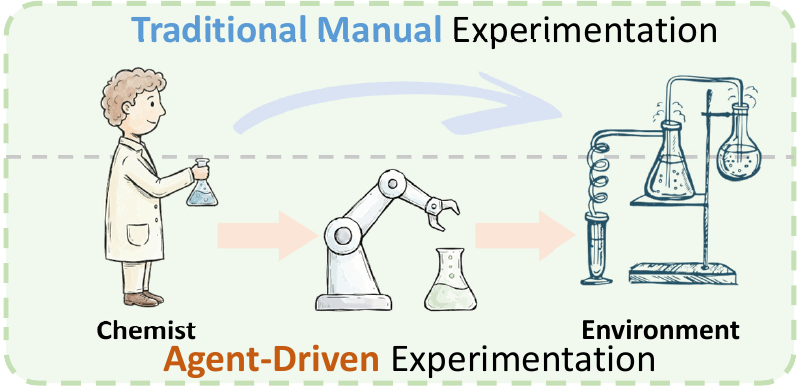}
    \caption{Paradigm shift from traditional manual to agent-driven experimentation via Embodied AI}
    \label{fig:structure-intro}
\end{figure}

Embodied AI is revolutionizing scientific discovery by enabling robots to automate hazardous and repetitive laboratory tasks \cite{jiang2015development, steiner2019chemputer, burger2020aichemist, zhu2022AI_Chemist}, as shown in Fig. \ref{fig:structure-intro}. However, achieving full autonomy in chemical laboratories remains challenging, as these experiments demand strict adherence to rigid, long-horizon, multi-stage protocols. The core bottleneck lies in grounding abstract natural language instructions into robust action sequences that account for dynamic environmental states. Recently, Vision-Language-Action (VLA) models have emerged as a powerful solution, leveraging the reasoning capabilities of Large Language Models (LLMs) \cite{team2025gemini, achiam2023gpt4} to bridge the gap between linguistic intent and physical execution \cite{black2024pi0, bjorck2025gr00t, shukor2025smolvla}.

Large Language Models (LLMs) have significantly advanced high-level planning for long-horizon robotic tasks \cite{wang2023voyager, darvish2025organa, song2025chemagent}. Extensive training equips these models with the ability to follow human instructions and perform complex reasoning \cite{guo2025deepseek}. Existing approaches, such as RoboMatrix \cite{mao2024robomatrix} and RoboChemist \cite{zhang2025robochemist}, represent milestones in skill-centric hierarchical planning and visual-semantic robot control. However, these frameworks often treat planning and execution as decoupled processes, struggling to maintain the rigorous, state-dependent closed-loop reasoning required for complex chemical protocols. Furthermore, they operate in an isolated manner, lacking mechanisms to consolidate successful execution patterns into a persistent, retrievable \textbf{long-term memory}. Consequently, the robot fails to learn from past experiences or accumulate knowledge across different experiments. While AI agents like OpenClaw \cite{openclaw2026} have demonstrated success in continuous, tool-use-based operation for desktop tasks \cite{tan2024cradle, zhang2025ufo2}, they fundamentally lack the embodied dexterity and physical grounding necessary to interact with the real-world laboratory environment.

\begin{figure*}[htbp]
    \centering
    \includegraphics[width=\linewidth]{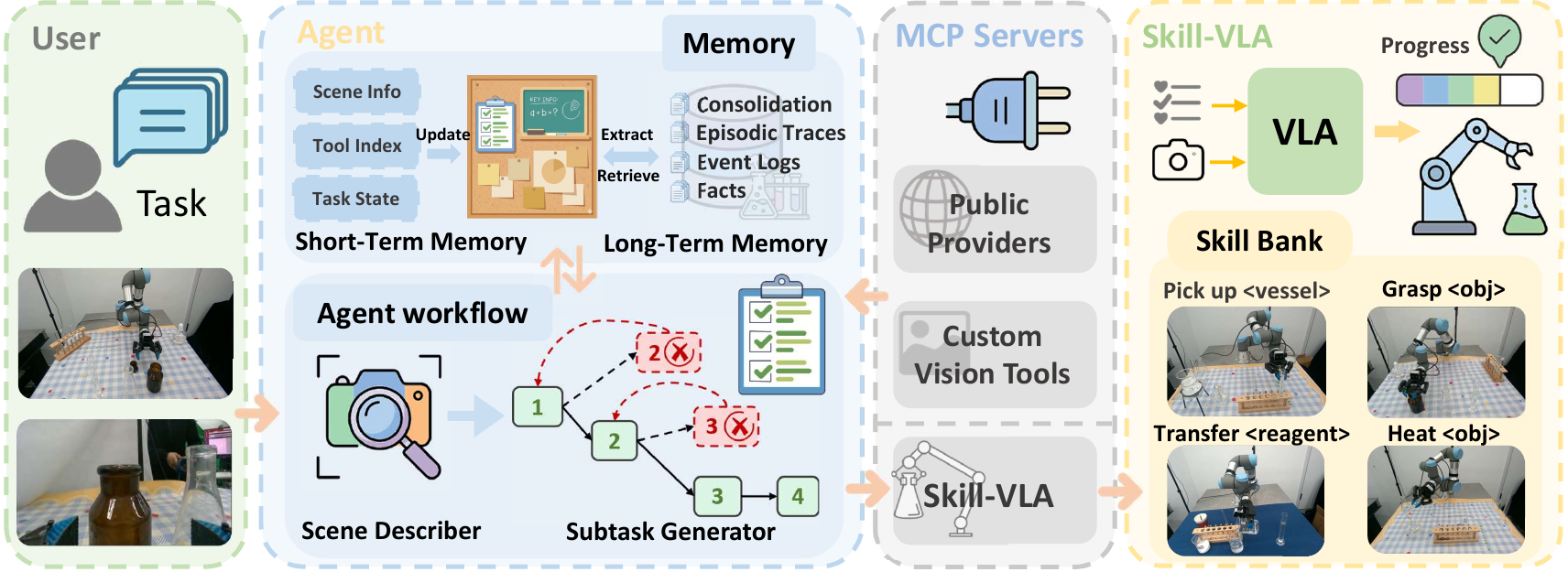}
    \caption{\textbf{Architecture of ChemBot}, a closed-loop framework for chemical task decomposition and action execution, centered on an AI agent integrating chain-backtracking subtask generation and dual-layer memory mechanisms, with a progress-aware Vision-Language-Action Model for low-level execution.}
    \vspace{-10pt}
    \label{fig:architecture}
\end{figure*}

To address these challenges, we introduce ChemBot, a closed-loop framework integrating an autonomous AI agent with a progress-aware VLA model (Skill-VLA), as illustrated in Fig.~\ref{fig:architecture}. Leveraging a Model Context Protocol (MCP) server for tool management, ChemBot features a dual-layer memory architecture: long-term memory consolidates successful trajectories as retrievable assets, while short-term memory structures real-time scene perception and task states. For execution, Skill-VLA generates continuous actions while simultaneously monitoring subtask progress. Finally, we implement a future-state-based asynchronous inference mechanism to mitigate trajectory discontinuities, ensuring operational safety and precision in complex chemical environments.

We demonstrate the efficacy of our framework through extensive real-world experiments on collaborative robots. The main contributions of this work are summarized as follows:

\begin{itemize}
\item \textbf{ChemBot Framework:} An agentic system for laboratory automation that integrates chain-backtracking task decomposition with a dual-layer memory mechanism, enabling robust, long-horizon reasoning.
\item \textbf{Skill-VLA Module:} An augmented VLA architecture featuring a progress-prediction head and a continuous inference pipeline, which enables real-time state monitoring and adaptation.
\item \textbf{Empirical Performance:} ChemBot outperforms VLA baselines in both decomposition and long-horizon execution, achieving a significantly higher overall success rate (SR).
\end{itemize}

\section{Related Work}
\textbf{LLM-based Embodied Agents.} 
Traditional robotic pipelines relying on hard-coded vision and control modules \cite{fang2023anygrasp} encounter difficulties in handling edge cases and generalizing across diverse environments. Consequently, large language models serve as high-level planners to decompose complex tasks into executable primitives \cite{ahn2022can, zhou2023language, zhu2023ghost, wang2023voyager}. To enhance reasoning efficiency, researchers employ search algorithms \cite{zhou2023language} optimized by world models \cite{hao2023reasoning} alongside cognitive prompting strategies for tool utilization \cite{xu2023creative}, problem abstraction \cite{zheng2023take}, and subgoal decomposition \cite{zhu2023ghost}. Wang et al. \cite{wang2024wonderful} proposed a multi-agent vision-language model framework for zero-shot robotic planning. Furthermore, Being-0 \cite{yuan2025being} adopts a dual-framework design comprising a foundation model for task planning and a connector combined with a skill library for execution. Ultimately, these components coalesce into comprehensive AI agent systems that integrate interactive planning and self-improving skill libraries \cite{wang2023voyager} to achieve robust physical exploration.

\textbf{Vision-Language-Action Models.} 
Recent advances in Vision-Language-Action (VLA) models, including  RT2 \cite{zitkovich2023rt_2}, $\pi_{0.5}$ \cite{intelligence2025pi05}, and GR00T \cite{bjorck2025gr00t}, leverage large-scale pre-training to establish general robot policies \cite{black2024pi0, kim2024openvla}. By integrating vision encoders with LLMs, these frameworks execute natural language instructions across diverse tasks and embodiments. To enhance robustness, CLAP \cite{li2025clap} incorporates a critic module for closed-loop reasoning, while RoboMatrix \cite{mao2024robomatrix} employs a skill-centric hierarchical structure for scalable task planning. Despite these strides, most VLAs still lack sophisticated reasoning for complex sequences. We address this by proposing an AI agent-based hierarchical system that enables autonomous planning and execution of chemical experiments.

\textbf{Autonomous Chemical Systems.} 
Chemical laboratory automation has transitioned from mechanical workstations executing basic operations \cite{jiang2015development} to autonomous platforms employing machine learning for standardized synthesis and material discovery \cite{steiner2019chemputer,burger2020aichemist,zhu2022AI_Chemist}. The recent integration of large language models and autonomous agents introduces natural language programming and advanced reasoning to chemical research. Systems like Co-scientist \cite{boiko2023autonomous} and CLARIFY \cite{yoshikawa2023clarify} convert human instructions into executable protocols while minimizing hallucinations. Improving physical execution, frameworks including ORGANA \cite{darvish2025organa} and RoboChemist \cite{zhang2025robochemist} utilize visual language models and task planning to manipulate complex materials safely. Finally, multiagent systems such as ChemAgents \cite{song2025chemagent} distribute literature review, experimental design, and robotic operation across specialized agents, enabling fully autonomous laboratory workflows.

\section{METHOD}
\begin{figure}[htbp]
    \centering
    \includegraphics[width=\linewidth]{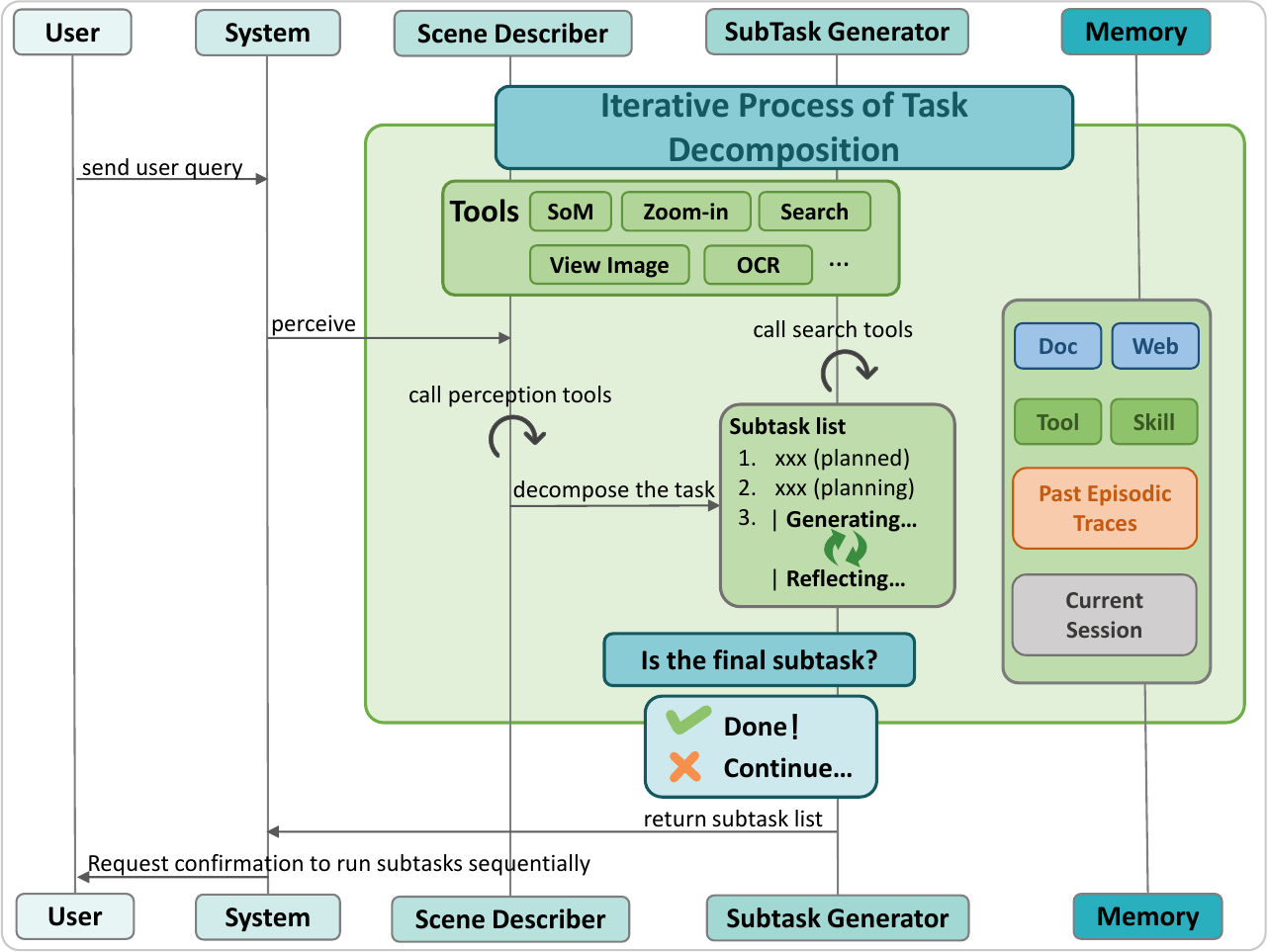}
    \vspace{-20pt}
    \caption{\textbf{Sequence diagram of iterative task decomposition via multi-agent collaboration}, illustrating a closed-loop workflow where an agent incrementally generates, validates, and refines subtasks through tool invocation and dual-layer memory retrieval until sequence completion.}
    \label{fig:framework-agent-workflow}
\end{figure}

\textbf{Overview.} To address the critical demand from ``AI for Science'' for automated chemistry experiments, we propose a hierarchical embodied intelligence architecture termed ``Agent-as-Planner, VLA-as-Skill''. The high-level layer governs global planning: by leveraging the comprehension capabilities of LLMs alongside specialized task-decomposition sub-agents and a dual-tier memory mechanism, it precisely deconstructs complex experimental protocols (the detailed workflow is illustrated in Fig.~\ref{fig:framework-agent-workflow}). For low-level execution, unlike small-scale, single-task models such as ACT \cite{zhao2023act}, our VLA module orchestrates a diverse set of motor skills within a unified network guided by natural language instructions. Ultimately, the VLA serves as a vital bridge, translating the logical reasoning of the AI agent into robust physical manipulations in the real world.

\subsection{Multi-Agent Incremental Task Planning}
\begin{figure}[htbp]
    \centering
    \vspace{-10pt}
    \includegraphics[width=\linewidth]{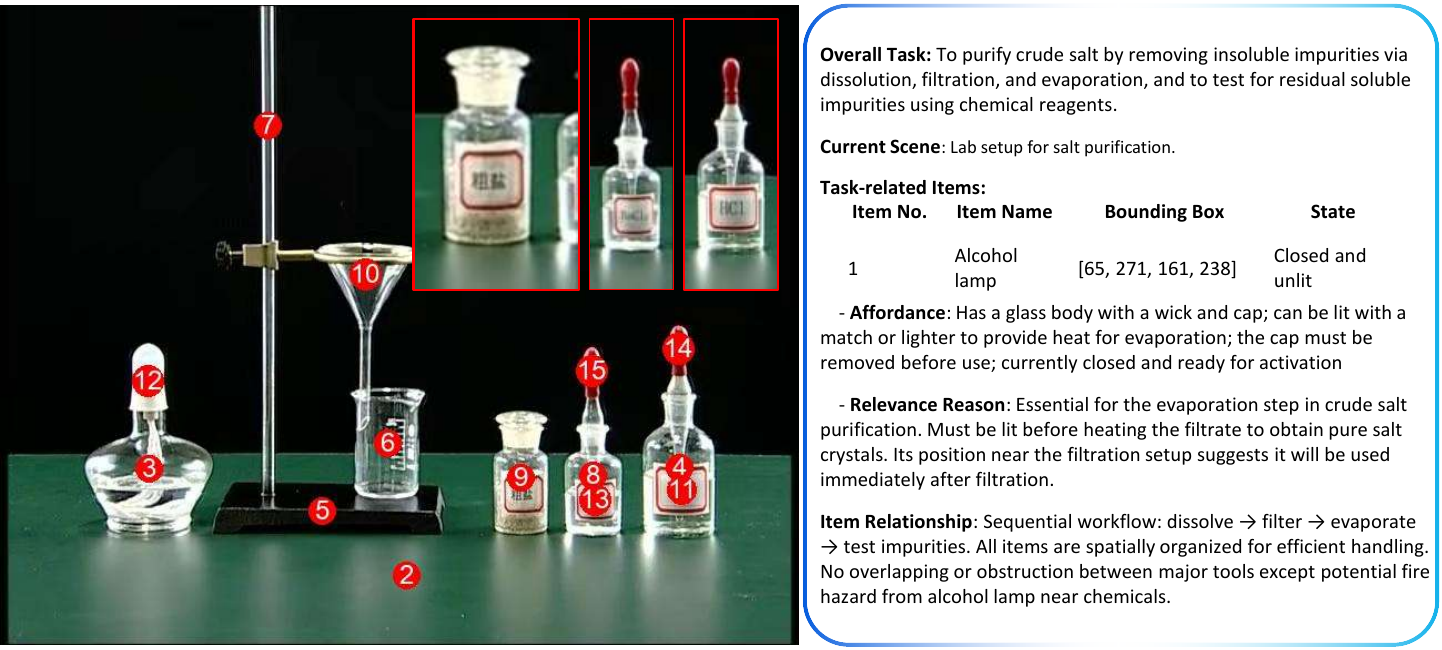}
    \vspace{-20pt}
    \caption{\textbf{Structured scene representation generated by the Scene Describer.} The module leverages Set-of-Mark prompting to parse raw lab images into a structured memory dashboard, detailing task-related items, their interactable states, affordances, and spatial relationships to provide visual grounding for subsequent task planning.}
    \label{fig:example-scene-describer}
\end{figure}

\textbf{Scene Describer}. Before planning, the agent must establish a precise structural comprehension of the current experimental environment. Therefore, we design the Scene Describer module, which autonomously invokes custom tools to parse the experimental bench image into a structured scene representation, providing a visual grounding foundation for subsequent incremental subtask planning, as shown in Fig. \ref{fig:example-scene-describer}. It initially employs Set-of-Mark Prompting \cite{yang2023som} to mark and enumerate objects within the image, identifying task-relevant equipment and reagents. The module then schedules visual assistance tools to systematically analyze object properties across multiple dimensions. It evaluates the interaction state of each object, such as container openness and occlusion. It assesses object affordances, including graspability, pourability, and available control mechanisms. The analysis additionally extracts spatial positions, collision constraints, causal connections, operational logic, and physical commonsense among the items.

\textbf{Subtask Generator}. The Subtask Reasoner and Reflector constitute an incremental planning and reflection validation loop. The Subtask Reasoner incorporates the Dashboard into the context, generating a single atomic subtask per iteration. The Reflector subsequently evaluates this subtask for feasibility, rationality, and non-redundancy. A valid subtask is appended to the execution queue, whereas an invalid one is marked for deletion, with feedback propagated back to the Reasoner to trigger backtracking and replanning. This iterative process continues until the Reflector determines that the task sequence is complete.

\subsection{Dual-Layer Memory Mechanism}

\textbf{Short-Term Working Memory.} Considering the capacity limitations and ephemeral storage characteristics of the model context, we propose a structured representation mechanism named Dashboard to maximize information utilization efficiency. This mechanism achieves context compression while dynamically tracking core states such as the experimental phase, current reagents, and pending objectives. The Dashboard operates through the collaboration of three modules. The scene information module serves as a structured representation of the scene generated by the Scene Describer. The tool index module converts massive information generated during tool utilization, such as images and retrieved content, into storage paths for lightweight representation, and it minimizes context window occupation through an on-demand loading strategy. Furthermore, the task state module continuously records the experimental planning progress to provide a macroscopic global perspective for incremental planning. The Dashboard adopts an immediate overwrite strategy where the system directly updates or appends specific field values upon environmental changes or task completion. 

\textbf{Episodic Long-Term Memory.} This component persistently preserves historical dialogues between experimenters and agents across distinct sessions. It enables semantic retrieval to leverage past episodic traces for new tasks, thereby facilitating the rapid planning and verification of novel experiments. Additionally, the system periodically extracts operational preference profiles and experimental archives from dialogue records to achieve long-term personalized adaptation.

\subsection{Progress-Aware VLA}
\begin{figure*}[htbp] 
    \centering
    \subfloat[Architecture of Skill-VLA]{
        \includegraphics[height=5.9cm]{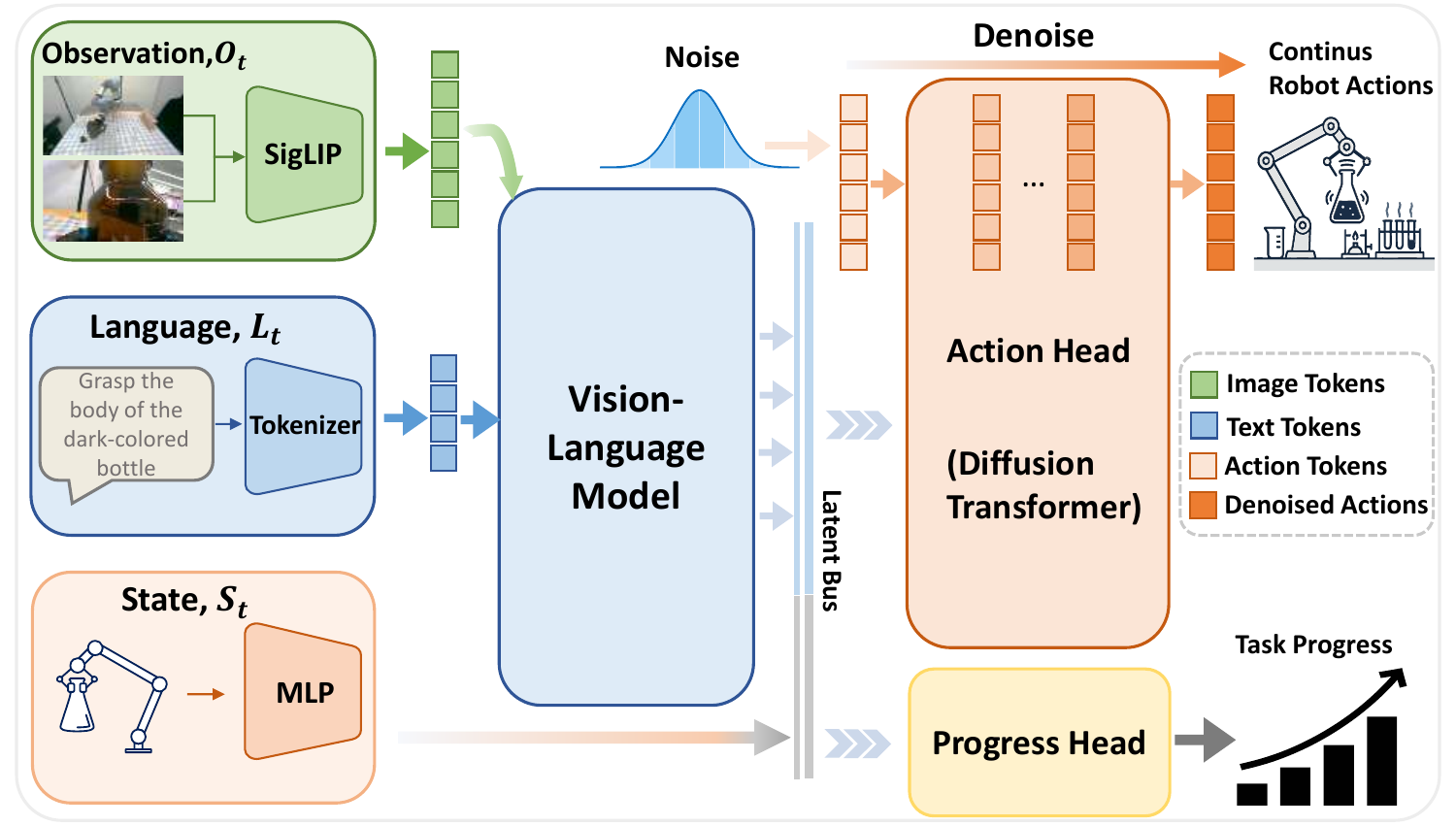}
        \label{fig:framework-VLA-progress}
    }
    \subfloat[Structure of the Progress Head]{
        \includegraphics[height=5.9cm]{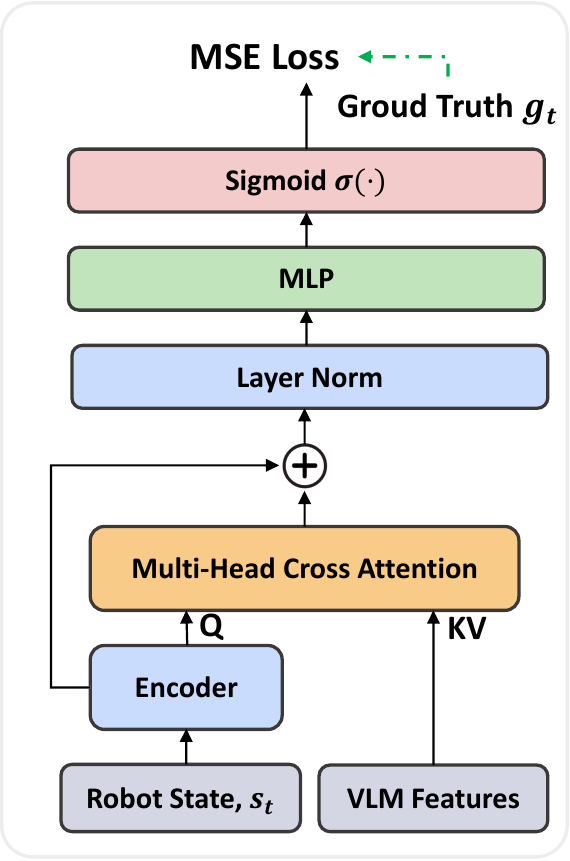}
        \label{fig:framework-progress-head}
    }
    \caption{\textbf{Detailed network architecture of Skill-VLA.} (a) The main VLA model for continuous action generation and subtask progress prediction. The state of the robotic arm and the hidden state of the VLM are provided as inputs to the Action Head and the Progress Head, respectively. (b) Internal structure of the Progress Head, an auxiliary module for the estimation of task progress.}
    \label{fig:VLA-framework-overall}
\end{figure*}

\textbf{VLA Policy.} 
We adopt GR00T \cite{bjorck2025gr00t} as our base VLA policy (Fig.~\ref{fig:VLA-framework-overall}). At environment step $t$, the model predicts an action chunk $A_t = [a_t, \dots, a_{t+H-1}]$, where $H$ is the action chunk size and each element $a_i$ denotes an individual robot action. This prediction is conditioned on context $c = (O_t, L_t, S_t)$ comprising visual observations, a language instruction, and the robot state. To generate $A_t$, a Diffusion Transformer (DiT) trained via Flow Matching maps Gaussian noise $x_0 \sim \mathcal{N}(0, I)$ to the target action distribution $x_1 = A_t$. Acting as a velocity estimator $v_\theta$ (with parameters $\theta$) conditioned on $c$, the DiT is optimized via:
$$\mathcal{L}_{flow} = \mathbb{E}_{\tau, x_0, x_1} \left[ \|v_\theta(x_\tau, \tau, c) - (x_1 - x_0)\|^2 \right]$$
where $\tau \in [0, 1]$ is the flow timestep and $x_\tau$ is the intermediate noisy state. During inference, $A_t$ is generated by iteratively denoising pure noise via standard Euler integration.

\textbf{Progress Head.} 
To estimate real-time subtask execution progress, we attach a lightweight neural network module to the VLA policy (Fig.~\ref{fig:framework-progress-head}). This module employs a cross-attention architecture: the robot state $S_t$ is linearly projected to serve as the query, while the final hidden representations from Eagle VLM (the VLM backbone of GR00T) act as the keys and values. The attention output is fused with the query via a residual connection and Layer Normalization, followed by a two-layer MLP with a Sigmoid activation to yield a continuous progress scalar $p_t \in [0, 1]$. 

This head is trained jointly with the VLA policy using a Mean Squared Error (MSE) loss, where ground-truth progress labels are computed as the normalized time-step position within the annotated subtask horizon. During inference, once the predicted progress $p_t$ exceeds a predefined completion threshold, the system signals the AI agent to terminate the current VLA skill and initiate the next subtask. This establishes a closed-loop, autonomous execution cycle without relying on human intervention or fixed time-step switching.

\subsection{Agent-as-Planner, VLA-as-Skill}

\textbf{Closed-Loop Chemical Manipulation.} To achieve seamless interoperability between the high-level planner and the low-level executor, we standardize the inference process of the VLA model into an Application Programming Interface (API) using the Model Context Protocol (MCP), an open-source standard connecting AI applications with external systems. Within this framework, the fine-tuned VLA model is encapsulated as a standardized atomic action tool. The tool receives disambiguated subtask instructions from the AI agent alongside continuous visual observations to drive the robotic arm in executing high-precision operations such as grasping and pouring. It synchronously returns binary execution states and diagnostic logs to facilitate closed-loop visual verification by the agent. The interface also integrates the previously described Progress Head mechanism to provide fine-grained execution state feedback. When the predicted execution progress exceeds a predefined threshold, the system automatically marks the current task as complete and triggers the agent to transition to the subsequent subtask. This dynamic termination mechanism circumvents the rigid reliance on fixed time steps typical of traditional control methods, thereby improving the smoothness and autonomy of long-horizon experimental operations.

\section{EXPERIMENT}

\subsection{Chemical Task Decomposition}

\textbf{Dataset.} Prior works, such as the CLARIFY dataset \cite{yoshikawa2023clarify}, provide a strong foundation by translating 108 chemical research and development (R\&D) protocols into the structured $\chi$DL format \cite{mehr2020universal}. We extend this text-based dataset by applying the Back-Instruct method \cite{slaughter2020task} to summarize overarching task objectives as inputs. Furthermore, we extract the requisite equipment and reagents to simulate scene perception, effectively compensating for the lack of visual information in text-only scenarios. To support multimodal inputs, we construct a vision-enhanced dataset using a standardized visual corpus of pedagogical demonstrations from foundational chemistry curricula. Following audio transcript extraction, we employ a collaborative pipeline that combines Vision-Language Model (VLM) assistance with human expert annotation to structure video keyframes and their corresponding execution steps. Ultimately, this pipeline yields 92 newly curated, high-quality, multimodal-aligned samples that serve as a reliable validation testbed for the AI agent's perception and reasoning capabilities.

\textbf{Evaluation Metrics.} To evaluate task decomposition, we employ three complementary metrics. First, the ROUGE family \cite{lin2004rouge} (ROUGE-1, ROUGE-2, and ROUGE-L) uses lexical overlap to measure word coverage and sequential logic matching between generated and reference steps. Second, BERTScore \cite{zhang2019bertscore} calculates cosine similarity via pre-trained contextual embeddings to capture synonym substitutions and sentence-level semantic consistency. Finally, to address the strong sequence dependency of chemical experiments, we propose an improved Levenshtein Distance (Edit Distance) \cite{yujian2007normalized}. It integrates step-level semantic alignment $s_{ij}$ with a positional penalty $\mathrm{Pos}(i, j) = \exp\left(-\frac{1}{\sigma} \left| \frac{i}{M-1} - \frac{j}{N-1} \right| \right)$ to ensure temporal consistency between the generated sequence of length $M$ and the reference sequence of length $N$. The normalized edit distance is defined as:
$$d_{\mathrm{edit}} = \frac{\sum_{i=1}^{M} (1 - \tilde{s}_i^*) + (N - |\mathcal{M}|)}{M}$$
where $\tilde{s}_i^* = s_{ij} \cdot \mathrm{Pos}(i, j)$ represents the position-weighted similarity of matched steps, and $|\mathcal{M}|$ is the total count of successful matches. This fine-grained quantification of omissions, deviations, and redundancies at the physical execution level successfully overcomes the limitations of traditional text metrics in assessing structural completeness.

\textbf{Performance Comparison of Base Models.} The visual and logical reasoning capabilities of the underlying base models directly determine the granularity and accuracy of chemical task decomposition. To this end, we evaluate three representative VLMs (Qwen3-VL-Flash \cite{yang2025qwen3}, Qwen3-VL-Plus, and Doubao-1.6 \cite{doubao1_6}) on complex chemical instructions. As detailed in Table~\ref{tab:main_results}, our evaluation framework spans three core dimensions: text generation quality (via ROUGE and BERTScore), step-level structural alignment (via our improved Edit Distance), and overall decomposition efficiency. Notably, the elevated execution latency observed for Doubao-1.6 is primarily attributed to cloud-based API response delays and network overhead during the evaluation period. Future work will involve local deployment of these models to obtain more deterministic inference speed metrics.

\begin{table*}[htbp]
\centering
\caption{\textbf{Performance comparison of different base models on chemical experiment task decomposition.} Evaluated metrics include normalized edit distance (Edit Dist.), BERTScore (P, R, F1), and ROUGE scores comparing generated task descriptions against ground-truth steps. Efficiency is assessed through total execution time, average time per step, and tool invocation count.}
\label{tab:main_results}
\resizebox{\textwidth}{!}{%
\begin{tabular}{l|cccccccccc}
\toprule
\multirow{2}{*}{\textbf{Model}} & \multirow{2}{*}{\textbf{Edit Dist.}$\downarrow$} & \multicolumn{3}{c}{\textbf{BERTScore}} & \multicolumn{3}{c}{\textbf{ROUGE}} & \multicolumn{3}{c}{\textbf{Efficiency}} \\
\cmidrule(lr){3-5} \cmidrule(lr){6-8} \cmidrule(lr){9-11}
 & & P$\uparrow$ & R$\uparrow$ & F1$\uparrow$ & R-1$\uparrow$ & R-2$\uparrow$ & R-L$\uparrow$ & Time (s)$\downarrow$ & Time/Step$\downarrow$ & Calls$\downarrow$ \\
\midrule
Qwen3-VL-Flash & 0.766 & \textbf{0.753} & 0.693 & 0.694 & 0.435 & 0.202 & 0.324 & \textbf{98.8} & 16 & 2.3 \\
Qwen3-VL-Plus   & \textbf{0.733} & 0.749 & 0.789 & \textbf{0.744} & \textbf{0.542} & \textbf{0.366} & \textbf{0.471} & 298.8 & 42 & 5.3 \\
Doubao-1.6     & 1.364 & 0.531 & \textbf{0.804} & 0.606 & 0.406 & 0.206 & 0.327 & 960.0 & 139 & 8.3 \\
\bottomrule
\end{tabular}%
}
\end{table*}

\textbf{Ablation Studies.} We designed a comprehensive set of ablation experiments (Table~\ref{tab:agent_ablation_settings}) to systematically isolate the contributions of each core component within ChemBot against the full-pipeline baseline (Setting 1). At the perception and planning level, removing the Scene Describer (Setting 2) eliminates initial visual analysis, demonstrating the necessity of multimodal scene understanding for accurate task initialization. Degrading the SubtaskAgent to a direct generation mode (Setting 3) isolates the impact of explicit chain reasoning on the decomposition accuracy of complex sequential tasks. To evaluate system robustness, eliminating the ReflectorAgent (Setting 4) disables self-correction, quantifying this module's critical role in preventing cascaded error diffusion. Finally, ablating the dual-layer memory mechanism (Setting 5) validates its efficacy in reducing context overhead and mitigating hallucinations during long-horizon operations.

\begin{table}[htbp]
\centering
\caption{Ablation settings of different modules in the ChemBot (base model: Qwen3-VL-Flash)}
\label{tab:agent_ablation_settings}
\renewcommand{\arraystretch}{1.2}
\resizebox{\linewidth}{!}{
    \begin{tabular}{c|ccccc}
    \toprule
    Setting No. & Describer & Subtask (Chain) & Reflector & Memory \\
    \midrule
    1 & \cmark & \cmark & \cmark & \cmark \\
    2 & \xmark & \cmark & \cmark & \cmark  \\
    3 & \cmark & \xmark & \cmark & \cmark  \\
    4 & \cmark & \cmark & \xmark & \cmark  \\
    5 & \cmark & \cmark & \cmark & \xmark  \\
    \bottomrule
    \end{tabular}
} 
\end{table}

\begin{table*}[htbp]
\centering
\caption{Ablation study of different modules in ChemBot (base model: Qwen3-VL-Flash)}
\label{tab:ablation_study}
\resizebox{\textwidth}{!}{%
\begin{tabular}{l|ccccccccc}
\toprule
\multirow{2}{*}{\textbf{Model Variant}} & \multirow{2}{*}{\textbf{Edit Dist.}$\downarrow$} & \multicolumn{3}{c}{\textbf{BERTScore}} & \multicolumn{3}{c}{\textbf{ROUGE Score}} & \multirow{2}{*}{\textbf{Time (s)}$\downarrow$} & \multirow{2}{*}{\textbf{Token Length}$\downarrow$} \\
\cmidrule(lr){3-5} \cmidrule(lr){6-8}
 & & P$\uparrow$ & R$\uparrow$ & F1$\uparrow$ & R-1$\uparrow$ & R-2$\uparrow$ & R-L$\uparrow$ & & \\
\midrule
Full Model & \textbf{0.766} & \textbf{0.753} & 0.693 & \textbf{0.694} & 0.435 & 0.202 & 0.324 & 98.8 & 22401 \\
w/o SceneDescriber & 0.915 & 0.680 & 0.510 & 0.556 & 0.365 & 0.143 & 0.268 & 63.6 & 19945 \\
w/o Subtask Chain & 1.195 & 0.573 & \textbf{0.825} & 0.646 & 0.412 & 0.201 & 0.301 & \textbf{43.1} & 21417 \\
w/o Reflector & 0.881 & 0.679 & 0.721 & 0.671 & 0.446 & \textbf{0.219} & \textbf{0.332} & 48.6 & 11686 \\
w/o Memory & 0.791 & 0.748 & 0.643 & 0.657 & \textbf{0.448} & 0.214 & 0.325 & 99.1 & 28064 \\
\bottomrule
\end{tabular}%
}
\end{table*}

We present the quantitative ablation results in Table~\ref{tab:ablation_study}, detailing the specific impact of each core module on overall task decomposition performance. As demonstrated by the degraded metrics across all variant settings, the complete ChemBot architecture remains indispensable for balancing reasoning accuracy and execution robustness in complex chemical scenarios.

\subsection{Real-World Evaluation}
We deploy the ChemBot system on a single-arm UR3 robotic platform to validate its real-world performance. These physical experiments are systematically designed to evaluate the system across three core dimensions: (1) Accuracy, defined as the success rate in completing precise chemical manipulation tasks; (2) Efficiency, quantified by the overall task execution time; and (3) Versatility, assessing the system's capability to robustly execute diverse combinations of chemical experimental procedures.


\textbf{Chemical Manipulation Dataset.} We present a multimodal dataset for chemical experiments, collected using a UR3 collaborative robot to support manipulation policy learning in real-world environments. We acquire data via a teleoperation paradigm utilizing VIVE controllers for precise end-effector pose mapping. Each trajectory comprises synchronized top-down and wrist-mounted video streams, alongside robot states $S_t$ and actions $a_t$. 

\begin{figure}[htbp]
    \centering
    \includegraphics[width=\linewidth]{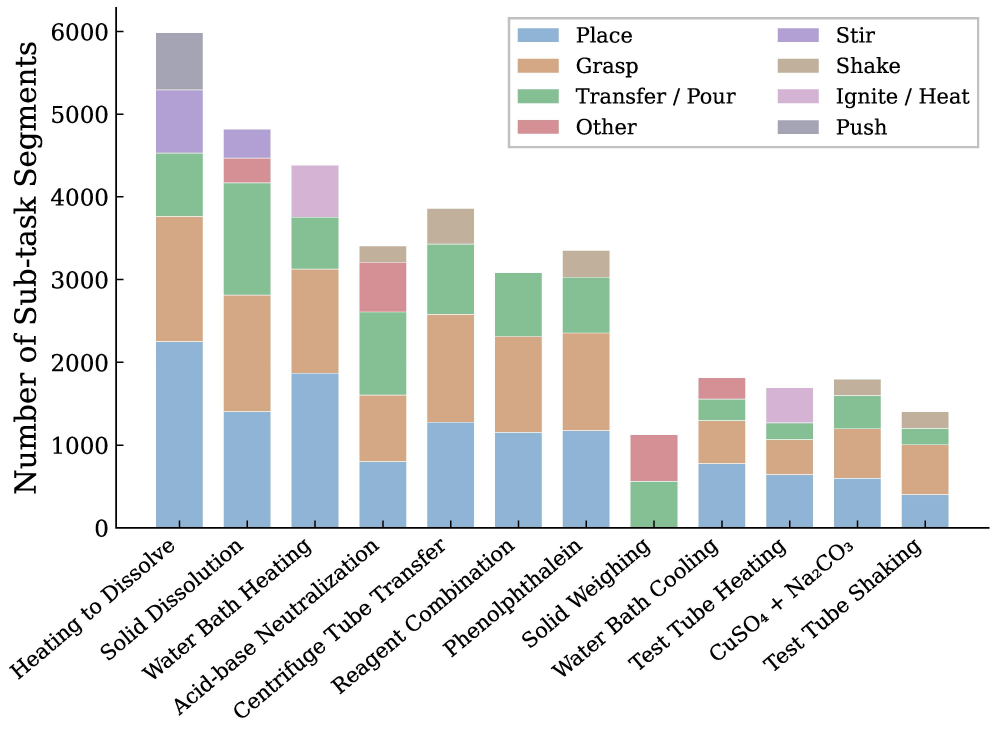}
    \vspace{-20pt}
    \caption{\textbf{Action composition of the top 12 chemical experiments.} This chart illustrates the quantitative distribution of fine-grained chemical manipulation primitives decomposed from complex expert trajectories to address long-horizon experimental challenges.}
    \label{fig:exp-action-per-experiment}
\end{figure}

To ensure high-quality supervision, we establish rigorous annotation guidelines by decomposing each experiment into predefined semantic subtasks. Annotators segment the top-down videos by recording the start and end frame indices for each subtask; trajectories failing to meet the complete specified sequence or containing execution errors are explicitly excluded from the training set. Within each subtask, fine-grained progress labels $p_t \in [0, 1]$ are generated via linear interpolation of frame indices. Our dataset encompasses 12 categories of foundational chemical procedures, such as test tube water bath heating and solid powder weighing. This process yields 5,459 expert trajectories (averaging 872.2 frames) and 51,580 atomic action segments, as detailed in Fig.~\ref{fig:exp-action-per-experiment}.

\textbf{Evaluation Metrics.} Traditional binary evaluation metrics \cite{shukor2025smolvla} often fail to capture the granular execution details and specific failure nodes inherent in long-horizon chemical tasks. To address this, we introduce a multi-level quantitative scoring mechanism and conduct 16 independent real-world trials for each task. The system assigns points based on a hierarchical success structure: (i) one point for each successful basic action (atomic success); (ii) one point for each completed logical action sequence (subtask success); and (iii) a final point for an error-free, end-to-end workflow (task success). The overall Success Rate (SR) is defined as $SR = (\sum S_{obs} / \sum S_{max}) \times 100\%$, where $S_{obs}$ denotes the accumulated score and $S_{max}$ is the theoretical maximum. 

\textbf{Experimental Setup.} All models are trained on the complete Chemical Manipulation Dataset. Since the full-trajectory baseline lacks the capability to autonomously maintain the execution sequence, we evaluate it within a 120s window per trial. In contrast, Skill-VLA achieves autonomous closed-loop execution via the progress head, requiring no manual intervention or time limits. For both approaches, the Success Rate (SR) is quantified by the number of successfully completed subtasks as verified by human experts. 

We evaluate Skill-VLA across three representative chemical experiments. The 11-step Precipitation task involves unsealing reagent bottles and transferring acid/base solutions via graduated cylinders into a flask for agitation. The 7-step Heat and Dissolution task requires precise manipulation of an alcohol lamp to facilitate the water bath heating of a test tube. Finally, the 6-step Neutralization task entails the controlled mixing of sodium carbonate and copper sulfate from separate test tubes.

\textbf{Implementation Details.} We fine-tune the Skill-VLA policy using the FluxVLA framework \cite{fluxvla2026}, which provides standardized training, inference, and real-robot deployment utilities for VLA models. The policy integrates the GR00T architecture with the proposed progress head and is trained on our Chemical Manipulation Dataset.

\begin{figure}[htbp]
    \centering
    \includegraphics[width=\linewidth]{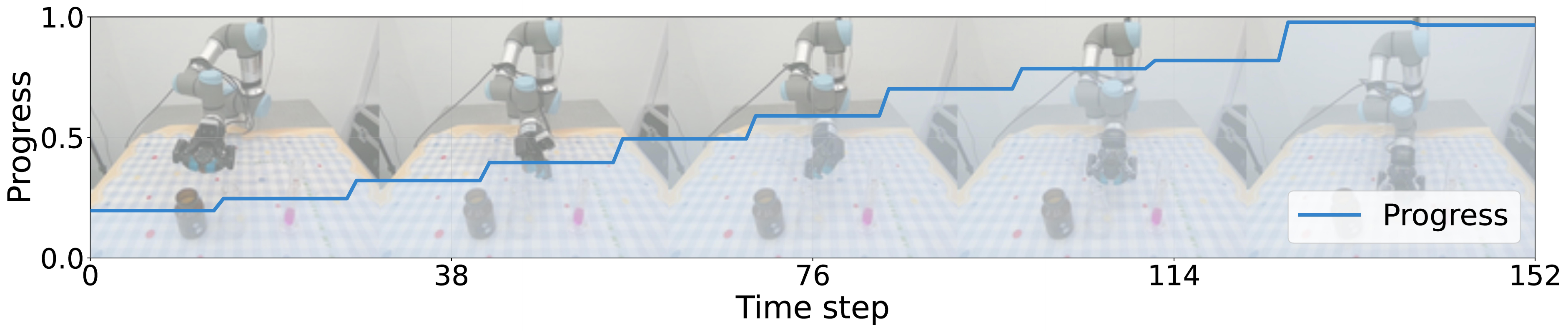}
    \vspace{-20pt}
    \caption{A visualization of the predicted task progress for opening a wide-mouth bottle.}
    \label{fig:exp-progress-with-scenes}
\end{figure}

\begin{figure}[htbp]
    \centering
    \includegraphics[width=\linewidth]{illustrations/scores_comparison.pdf}
    \vspace{-20pt}
    \caption{\textbf{Bar chart of success rates for three chemistry tasks} (error bars show 68\% Wilson score intervals). Skill-VLA with subtask-level training outperforms the full-trajectory baselines ($\pi_{0.5}$, GR00T) across all tasks, validating its advantage in long-horizon tasks.}
    \label{fig:exp-async-infer-SR}
\end{figure}

\textbf{Results.} 
The results in Fig. \ref{fig:exp-async-infer-SR} demonstrate that our complete strategy of integrating subtask decomposition with progress prediction (shown in Fig. \ref{fig:exp-progress-with-scenes}) achieves significantly higher success rates across all evaluated tasks compared to alternative baselines. The full-trajectory baseline suffers from severe distribution shifts over long horizons and compounding errors. It frequently drifts from the intended trajectory during the intermediate and terminal stages of the tasks, which yields the lowest overall performance. Constrained by the memoryless nature of standard VLA models, this baseline is incapable of implicitly tracking task progress. For instance, given largely identical inputs, repetitive grasping and releasing actions frequently occur, causing the system to skip intermediate steps. In contrast, the incorporation of progress prediction empowers the system to dynamically assess subtask completion in real time utilizing multimodal features, thereby effectively eliminating the reliance on fixed execution steps. Ultimately, this autonomous closed loop of planning, execution, and perception substantially mitigates the inherent deficiencies in basic execution.


\subsection{Asynchronous Continuous Inference}
\textbf{Asynchronous Inference.} To prevent hazardous control pauses and potential chemical spills, we decouple model inference from the control loop to enable asynchronous execution. Given an inference delay $d$, an execution horizon $s$, and an action horizon $H$ ($d \le s < H$), a cycle initiated at step $t$ concludes at $t+d$, where we proactively truncate the first $d$ expired actions from the new sequence $A_{new}$ and overwrite the remaining unexecuted portion of $A_{old}$. To mitigate trajectory jumps during this latency, we employ Training-time RTC \cite{black2025training} by fixing the flow matching timestep to 1.0 for the initial $d$ actions, effectively masking the prefix and focusing denoising optimization on the sequence suffix. Building upon this, we incorporate the predicted future state $s_{t+d}$ as the boundary constraint and decision anchor instead of the current state $s_t$, ensuring that generated actions align precisely with the robot's physical configuration at the moment of execution to maintain spatial continuity and smooth asynchronous control.

\begin{figure}[htbp]
    \centering
    \includegraphics[width=\linewidth]{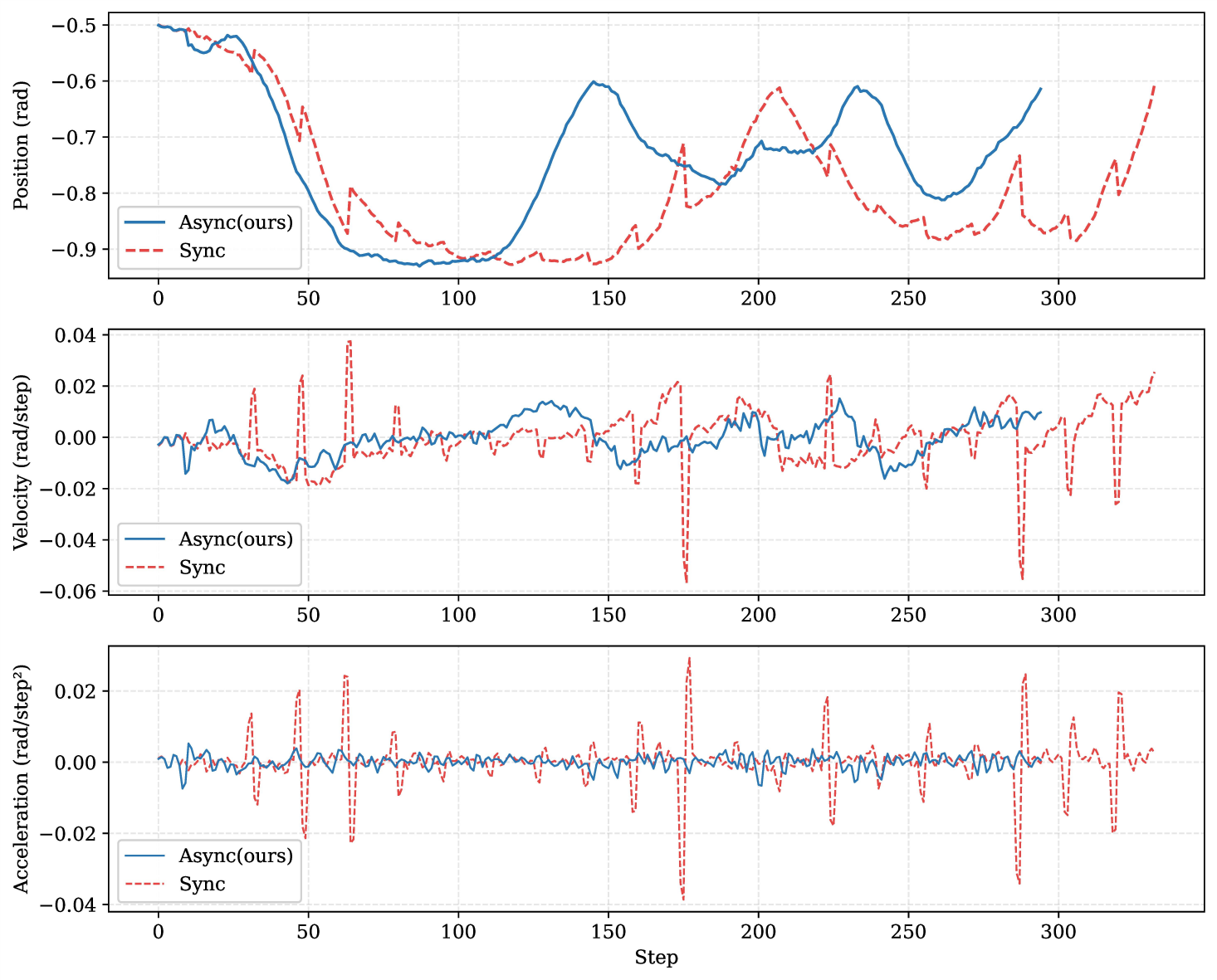}
    \vspace{-20pt}
    \caption{Comparison of motion trajectories for Joint 4 of the UR robot between synchronous and asynchronous inference. Asynchronous inference results in smoother trajectories and reduced execution time.}
    \label{fig:exp-joint-traj-comparison}
\end{figure}

As illustrated in Fig.~\ref{fig:exp-joint-traj-comparison}, the optimized trajectories exhibit substantially greater smoothness compared to those of the baseline. In real-world deployments, such characteristics effectively reduce end-effector jitter and enable continuous motion execution with minimal fluctuations.

\section{CONCLUSIONS}

This paper presented ChemBot, a hierarchical framework that decouples high-level reasoning from low-level execution to advance autonomous chemical experimentation. By integrating MCP-based skill encapsulation, a progress-aware inference pipeline, and asynchronous closed-loop control, ChemBot achieves superior robustness compared to end-to-end VLA systems. Despite these advancements, limitations persist regarding LLM-driven planning reliability, visual perception of transparent laboratory glassware, and generalization across novel scenes. Future work will address these constraints by scaling to cross-device manipulation and enhancing agentic reasoning, ultimately aiming to realize fully autonomous scientific assistants capable of end-to-end experiment orchestration and analysis.









\bibliographystyle{IEEEtran} 
\bibliography{cite}

\end{document}